\documentclass{article}
\usepackage{enumitem}

\PassOptionsToPackage{prologue,dvipsnames}{xcolor}

\usepackage{adjustbox}
\usepackage{booktabs}
\usepackage{multirow}
\usepackage{soul}

\usepackage[table]{xcolor}

\usepackage{pgfplots}
\usepackage{url}
\usepackage[most]{tcolorbox}
\usepackage{multicol}

\usepackage{verbatim}
\usepackage{amssymb}
\usepackage{pifont}
\usepackage{mathrsfs}
\usepackage{siunitx} 

\definecolor{codegreen}{rgb}{0,0.6,0}
\definecolor{codegray}{rgb}{0.5,0.5,0.5}
\definecolor{codepurple}{rgb}{0.58,0,0.82}
\definecolor{backcolour}{rgb}{0.95,0.95,0.92}
\definecolor{wkgreen}{RGB}{184,244,175}
\definecolor{wkpurple}{RGB}{210,210,253}
\definecolor{wkyellow}{RGB}{255,241,177}
\definecolor{wkblue}{RGB}{174,217,253}
\usepackage{listings}

\lstdefinestyle{mystyle}{
  commentstyle=\color{codegreen},
  keywordstyle=\color{magenta},
  numberstyle=\tiny\color{codegray},
  stringstyle=\color{codepurple},
  basicstyle=\ttfamily\tiny,
  breakatwhitespace=false,         
  breaklines=true,                 
  captionpos=b,                    
  keepspaces=false,                                 
  showspaces=false,                
  showstringspaces=false,
  showtabs=false,                  
  tabsize=2
}
\usepackage[preprint]{neurips_2025}

\usepackage{xspace}
\newcommand{\model}{\textsc{R2P}\xspace}

\usepackage[utf8]{inputenc} 
\usepackage[T1]{fontenc}    
\usepackage{hyperref}       
\usepackage{url}            
\usepackage{booktabs}       
\usepackage{amsfonts}       
\usepackage{nicefrac}       
\usepackage{microtype}      
\usepackage{xcolor}         
\usepackage{wrapfig}

\usepackage{multirow} 
\usepackage{geometry}

\usepackage{graphicx}
\usepackage{stackengine}

\title{Reasoning Meets Personalization: \\Unleashing the Potential of Large Reasoning Model\\ for Personalized Generation}

\author{%
  Sichun Luo\textsuperscript{1,2} \quad Guanzhi Deng\textsuperscript{2} \quad Jian Xu\textsuperscript{3} \quad Zerui Yang\textsuperscript{2} \\ \textbf{Xiaojie Zhang\textsuperscript{4} \quad Hanxu Hou\textsuperscript{1*} \quad Linqi Song\textsuperscript{2}\thanks{Corresponding Author}}
    \\\\
  \textsuperscript{1}Dongguan University of Technology \quad \textsuperscript{2}City University of Hong Kong \\
  \textsuperscript{3}Tsinghua University \quad
  \textsuperscript{4}Guangzhou University \\
  \texttt{sichunluo2@gmail.com} \\
}

\begin{document}

\maketitle

\begin{abstract}
Personalization is a critical task in modern intelligent systems, with applications spanning diverse domains, including interactions with large language models (LLMs). Recent advances in reasoning capabilities have significantly enhanced LLMs, enabling unprecedented performance in tasks such as mathematics and coding. However, their potential for personalization tasks remains underexplored. 

In this paper, we present the first systematic evaluation of large reasoning models (LRMs) for personalization tasks. Surprisingly, despite generating more tokens, LRMs do not consistently outperform general-purpose LLMs, especially in retrieval-intensive scenarios where their advantages diminish. Our analysis identifies three key limitations: divergent thinking, misalignment of response formats, and ineffective use of retrieved information. To address these challenges, we propose \underline{R}einforced \underline{R}easoning for \underline{P}ersonalization ({\model}), a novel framework that incorporates a hierarchical reasoning thought template to guide LRMs in generating structured outputs. In addition, we introduce an intervention method in the reasoning process to enforce the adherence to the designed reasoning patterns to enhance alignment. We also propose a self referencing mechanism to ensure consistency. Extensive experiments demonstrate that our approach significantly outperforms existing techniques.

\end{abstract}

\section{Introduction}

Personalization is a cornerstone of modern artificial intelligence, driving transformative advancements in applications such as recommendation systems \cite{luo2025recranker,luo2022personalized,luo2024perfedrec++}, adaptive user interfaces \cite{jameson2007adaptive,schiaffino2004user}, and conversational agents \cite{kocaballi2019personalization,araujo2024speaking,luo2025clarifymt}. 
At its core, personalization demands the ability of both deep contextual understanding and nuanced reasoning. 
Large language models (LLMs) \cite{brown2020language,anil2023palm,achiam2023gpt,jiang2024mixtral,touvron2023llama,touvron2023llama2} have emerged as powerful tools for this purpose, leveraging their capacity to generate human-like text and process complex linguistic patterns. Recent strides in reasoning capabilities have further elevated LLMs, enabling them to excel in structured tasks such as mathematical problem solving \cite{wangmathcoder,zhousolving}, code generation \cite{gu2023llm,liu2024exploring,luo2026alphaqt}, algorithm discovery \cite{luo2026relay,luo2026seaevo}, and logical inference \cite{wei2022chain,yao2023tree}. These advancements have fueled expectations that large reasoning models (henceforth referred to as LRMs) could revolutionize personalization by delivering outputs finely attuned to individual user needs. Yet, despite their promise, the efficiency of LRMs in personalization tasks remains largely unexplored, leaving critical questions about their strengths and limitations unanswered.
This raises a natural question: \textit{can LRMs surpass LLMs in the personalization task?}


In this work, we present the first systematic evaluation of LRMs for personalization tasks, focusing on their performance across a diverse set of tasks defined by the Language Model Personalization (LaMP) benchmark \cite{salemi2024lamp}. 
LaMP is a widely used dataset for personalization evaluation.
Our findings reveal a surprising and counterintuitive trend: despite their superior reasoning abilities, LRMs do not consistently outperform general-purpose LLMs in personalization tasks. This performance gap becomes particularly pronounced when retrieval-augmented generation (RAG) \cite{lewis2020retrieval} is employed, suggesting that LRMs struggle to effectively leverage external knowledge for user-specific adaptation.

To understand this phenomenon, our detailed analysis reveals three key limitations of LRMs in personalization tasks. First, LRMs excel in convergent reasoning for well-defined problems like equation solving but lack the divergent thinking needed to capture nuanced user preferences, often yielding sub-optimal outputs. Second, they struggle to produce consistent, structured responses aligned with required formats. Finally, LRMs inefficiently leverage retrieved context in RAG settings, prioritizing internal logic over user-specific information, which hampers effective personalization.

\begin{figure*}
  \centering
\includegraphics[width=\textwidth]{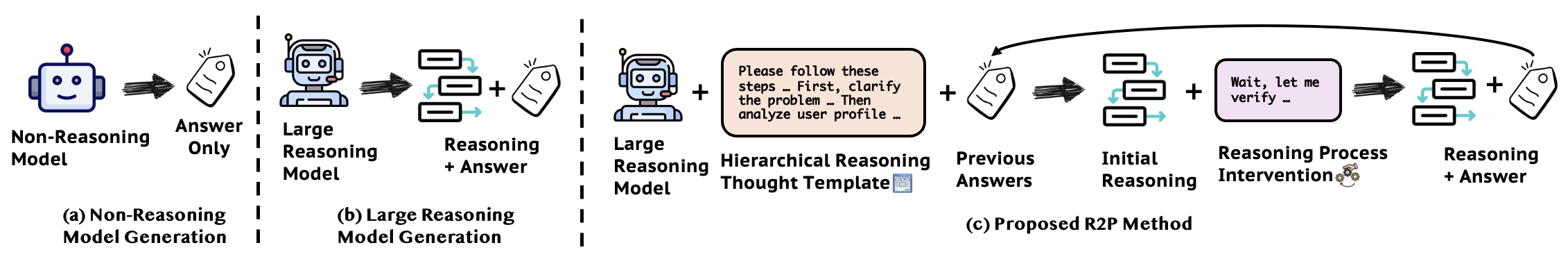}
    \caption{Comparison between non-reasoning model generation, large reasoning model generation, and proposed \model method.}
\label{fig:example}
\end{figure*}



To address these limitations, we propose \textbf{R}einforced \textbf{R}easoning for \textbf{P}ersonalization (\textbf{\model}), a novel training-free framework that enhances LRMs for personalization tasks. \model employs a hierarchical Reasoning thought template that systematically decomposes tasks and integrates user-specific context to guide structured reasoning generation, ensuring effective use of retrieved data.
It incorporates dynamic intervention to monitor and correct reasoning deviations, enforcing alignment with task specifications and user preferences. 
Additionally, consistency checks synthesize multiple outputs to ensure coherent and personalized results, collectively enhancing divergent thinking, format consistency, and retrieval integration for consistent user-tailored results.
An example comparing non-reasoning model generation, large reasoning model generation, and the proposed R2P method is shown in Figure \ref{fig:example}.

In a nutshell, our contributions are threefold:

\begin{itemize}[left=0pt]
    \item We perform a comprehensive evaluation of general-purpose LLMs and LRMs on the LaMP dataset, uncovering interesting insights into their strengths and limitations for personalization tasks, followed by detailed analysis.
    \item We introduce \model, a novel framework that incorporates a hierarchical reasoning thought template to guide LRMs in producing structured outputs. Additionally, we propose a reasoning process intervention method to enforce adherence to designed reasoning patterns and a self referencing mechanism to ensure output consistency.

    \item Through extensive experiments, we demonstrate that \model achieves superior performance on personalization tasks, outperforming baseline methods.
\end{itemize}

\section{Related Work}

\textbf{LLM Personalization.}
Personalizing large language models (LLMs) has gained significant attention as a means to tailor responses to individual user preferences and behaviors \cite{chen2024large,zhang2024personalization,zhuanghydra}. Existing research primarily focuses on designing prompts that incorporate user-generated content, historical interactions, or behavioral patterns to enhance personalization. 
Early work in LLM personalization leverages in-context learning and few-shot prompting to encode user behavior or preferences directly into the input \cite{luo2024integrating}. To address the challenge of growing user data, retrieval-augmented personalization has emerged as a scalable solution \cite{huang2023learning,salemi2024lamp}. This approach retrieves the most relevant records from a user’s behavioral history to augment prompts efficiently. Another line of research summarizes user preferences and behaviors into concise natural language profiles for prompt augmentation \cite{richardson2023integrating}.

\textbf{Large Reasoning Model.}
The quest to enhance reasoning in LLMs has led to the emergence of Large Reasoning Models (LRMs), designed to tackle complex tasks requiring multi-step inference, logical reasoning, and contextual understanding \cite{huang2023towards,xu2025towards}. 
A foundational approach to improving reasoning is the development of prompting techniques that guide models toward structured thinking. Chain-of-Thought (CoT) prompting \cite{wei2022chain} encourages step-by-step reasoning, while extensions like Tree-of-Thoughts (ToT) \cite{yao2023tree} explore multiple reasoning paths to enhance solution robustness. These methods have been proven effective across tasks such as arithmetic reasoning and commonsense inference. However, their reliance on carefully crafted prompts can limit generalizability, particularly for novel or underspecified tasks.
To address this, researchers have explored fine-tuning LLMs on specialized reasoning datasets for advanced mathematical problems \cite{wangmathcoder}. 
Recent models exemplify the LRM paradigm. OpenAI’s o1 \cite{jaech2024openai} demonstrates strong performance in mathematical, logical, and programming tasks by integrating specialized reasoning modules and extended inference-time computation. Similarly,  Deepseek R1 \cite{guo2025deepseek} leverage optimized architectures and training objectives to prioritize reasoning over rote memorization. These models highlight a trend toward balancing scale with reasoning-specific design, though challenges remain in ensuring robustness across diverse domains and mitigating computational overhead.

\textbf{Limitation.} 
While LRMs demonstrate strong performance in tasks such as mathematical and logical reasoning, their effectiveness in personalization tasks remains underexplored, leaving a significant research gap. To address this, we provide the first systematic evaluation of LRMs compared to general-purpose LLMs on the LaMP benchmark, shedding light on their capabilities and limitations in adapting to user-specific contexts.

\begin{table}[t]
\centering
\vspace{-0.1in}
\caption{Performance comparison between Llama3 and DeepSeek-Llama3 on LaMP tasks. The best results are highlighted in \textbf{bold}.}
\label{tab:lamp_results1}
\resizebox{\textwidth}{!}{
\begin{tabular}{l l c c c c c c c c}
\toprule
\multirow{3}{*}{Task} & \multirow{3}{*}{Metric} & \multicolumn{2}{c}{Non-Personalized (k=0)} & \multicolumn{2}{c}{Random (k=1)} & \multicolumn{2}{c}{RAG (k=1)} & \multicolumn{2}{c}{RAG (k=4)} \\
\cmidrule(r){3-4} \cmidrule(lr){5-6} \cmidrule(lr){7-8} \cmidrule(l){9-10}
& & Llama3 & \stackanchor{DeepSeek-}{Llama3} & Llama3 & \stackanchor{DeepSeek-}{Llama3} & Llama3 & \stackanchor{DeepSeek-}{Llama3} & Llama3 & \stackanchor{DeepSeek-}{Llama3} \\
\midrule
LaMP-1 & Acc $\uparrow$ & \textbf{0.490} & 0.465 & \textbf{0.640} & 0.625 & \textbf{0.675} & 0.650 & \textbf{0.760} & 0.712 \\
       & F1 $\uparrow$  & \textbf{0.480} & 0.461 & \textbf{0.639} & 0.625 & \textbf{0.675} & 0.649 & \textbf{0.760} & 0.711 \\
\midrule
LaMP-2N & Acc $\uparrow$ & \textbf{0.615} & 0.605 & 0.670 & \textbf{0.687} & 0.640 & \textbf{0.698} & \textbf{0.795} & 0.697 \\
       & F1 $\uparrow$  & 0.290 & \textbf{0.336} & 0.341 & \textbf{0.393} & 0.315 & \textbf{0.421} & \textbf{0.497} & 0.392 \\
\midrule
LaMP-2M & Acc $\uparrow$ & \textbf{0.415} & 0.350 & \textbf{0.310} & 0.295 & 0.320 & \textbf{0.332} & \textbf{0.455} & 0.403 \\
       & F1 $\uparrow$  & \textbf{0.265} & 0.228 & \textbf{0.224} & 0.218 & 0.199 & \textbf{0.257} & \textbf{0.343} & 0.287 \\
\midrule
LaMP-3 & MAE $\downarrow$ & \textbf{0.445} & 0.467 & 0.495 & \textbf{0.448} & \textbf{0.360} & 0.383 & \textbf{0.280} & 0.325 \\
       & RMSE $\downarrow$ & \textbf{0.745} & 0.751 & 0.919 & \textbf{0.767} & 0.748 & \textbf{0.678} & \textbf{0.656} & 0.664 \\
\midrule
LaMP-4 & R-1 $\uparrow$ & 0.093 & \textbf{0.149} & 0.123 & \textbf{0.142} & 0.125 & \textbf{0.152} & \textbf{0.179} & 0.166 \\
       & R-L $\uparrow$ & 0.083 & \textbf{0.131} & 0.108 & \textbf{0.125} & 0.110 & \textbf{0.133} & \textbf{0.157} & 0.148 \\
\midrule
LaMP-5 & R-1 $\uparrow$ & 0.114 & \textbf{0.381} & 0.130 & \textbf{0.389} & 0.148 & \textbf{0.406} & 0.304 & \textbf{0.413} \\
       & R-L $\uparrow$ & 0.106 & \textbf{0.308} & 0.114 & \textbf{0.314} & 0.132 & \textbf{0.341} & 0.272 & \textbf{0.344} \\
\midrule
LaMP-7 & R-1 $\uparrow$ & 0.351 & \textbf{0.426} & \textbf{0.379} & 0.318 & \textbf{0.370} & 0.338 & \textbf{0.415} & \textbf{0.415} \\
       & R-L $\uparrow$ & 0.291 & \textbf{0.370} & \textbf{0.325} & 0.272 & \textbf{0.320} & 0.293 & \textbf{0.362} & \textbf{0.362} \\

\bottomrule
\end{tabular}}
\end{table}

\begin{table}[t]
\centering
\vspace{-0.1in}
\caption{Average token length comparison of Llama3 and DeepSeek-Llama3 on LaMP tasks.}
\label{tab:lamp_length_results}
\resizebox{\textwidth}{!}{
\begin{tabular}{l c c c c c c c c}
\toprule
\multirow{3}{*}{Task} & \multicolumn{2}{c}{Non-Personalized (k=0)} & \multicolumn{2}{c}{Random (k=1)} & \multicolumn{2}{c}{RAG (k=1)} & \multicolumn{2}{c}{RAG (k=4)} \\
\cmidrule(r){2-3} \cmidrule(lr){4-5} \cmidrule(lr){6-7} \cmidrule(l){8-9}
& Llama3 & \stackanchor{DeepSeek-}{Llama3} & Llama3 & \stackanchor{DeepSeek-}{Llama3} & Llama3 & \stackanchor{DeepSeek-}{Llama3} & Llama3 & \stackanchor{DeepSeek-}{Llama3} \\ \midrule
LaMP-1 & 3.00 & 523.40 & 3.00 & 721.57 & 3.00 & 671.59 & 3.00 & 1007.78 \\
LaMP-2N & 1.87 & 317.02 & 1.85 & 327.33 & 1.85 & 318.17 & 1.86 & 315.60 \\
LaMP-2M & 7.76 & 473.67 & 2.53 & 468.70 & 2.37 & 476.00 & 2.00 & 437.53 \\
LaMP-3 & 1.00 & 347.90 & 1.12 & 365.36 & 1.00 & 355.40 & 1.00 & 350.80 \\
LaMP-4 & 99.81 & 436.96 & 65.62 & 446.96 & 64.80 & 450.77 & 22.68 & 460.37 \\
LaMP-5 & 168.48 & 426.67 & 151.59 & 469.33 & 145.28 & 453.20 & 94.22 & 548.12 \\
LaMP-7 & 21.22 & 307.84 & 22.12 & 395.40 & 22.66 & 395.74 & 25.90 & 386.19 \\

\midrule
Avg. & 43.31 & 404.78 & 35.40 & 456.38 & 34.42 & 445.84 & 21.52 & 500.91 \\
\bottomrule
\end{tabular}}
\end{table}

\begin{table}[h]
\centering
\caption{Performance comparison between Qwen and DeepSeek-Qwen on LaMP tasks with RAG ($k=1$). The best results are highlighted in \textbf{bold}. }
\label{tab:lamp_results2}
\resizebox{\textwidth}{!}{
\begin{tabular}{l l c c c c c c c c}
\toprule
Task & Metric & Qwen1.5 & Qwen7 & Qwen14 & Qwen32 & \stackanchor{DeepSeek-}{Qwen1.5} & \stackanchor{DeepSeek-}{Qwen7} & \stackanchor{DeepSeek-}{Qwen14} & \stackanchor{DeepSeek-}{Qwen32} \\
\midrule
LaMP-1 & Acc $\uparrow$ & 0.615 & \textbf{\underline{0.710}} & 0.670 & 0.680 & 0.593 & 0.642 & 0.665 & \underline{0.670} \\
       & F1 $\uparrow$  & 0.576 & \textbf{\underline{0.706}} & 0.669 & 0.674 & 0.593 & 0.640 & 0.664 & \underline{0.668} \\
\midrule
LaMP-2N & Acc $\uparrow$ & 0.645 & 0.690 & 0.675 & \underline{0.720} & 0.578 & 0.650 & 0.700 & \textbf{\underline{0.720}} \\
       & F1 $\uparrow$  & 0.299 & 0.384 & 0.374 & \underline{0.422} & 0.291 & 0.348 & \textbf{\underline{0.426}} & 0.424 \\
\midrule
LaMP-2M & Acc $\uparrow$ & 0.245 & 0.315 & 0.360 & \underline{0.410} & 0.213 & 0.280 & \textbf{\underline{0.443}} & 0.438 \\
       & F1 $\uparrow$  & 0.188 & 0.260 & 0.274 & \underline{0.296} & 0.143 & 0.227 & 0.349 & \textbf{\underline{0.355}} \\
\midrule
LaMP-3 & MAE $\downarrow$ & 0.525 & 0.560 & 0.380 & \underline{0.355} & 0.690 & 0.455 & \textbf{\underline{0.280}} & 0.328 \\
       & RMSE $\downarrow$ & 0.875 & 0.870 & 0.510 & \underline{0.485} & 1.190 & 0.672 & \textbf{\underline{0.330}} & 0.378 \\
\midrule
LaMP-4 & R-1 $\uparrow$ & \underline{0.153} & 0.137 & 0.151 & 0.136 & 0.129 & 0.147 & \textbf{\underline{0.162}} & 0.157 \\
       & R-L $\uparrow$ & \underline{0.134} & 0.126 & 0.132 & 0.117 & 0.115 & 0.131 & \textbf{\underline{0.142}} & 0.138 \\
\midrule
LaMP-5 & R-1 $\uparrow$ & \textbf{\underline{0.455}} & 0.444 & 0.395 & 0.304 & 0.363 & 0.414 & \underline{0.444} & 0.435 \\
       & R-L $\uparrow$ & \textbf{\underline{0.385}} & 0.374 & 0.329 & 0.264 & 0.302 & 0.344 & \underline{0.376} & 0.364 \\
\midrule
LaMP-7 & R-1 $\uparrow$ & 0.381 & \textbf{\underline{0.461}} & 0.460 & 0.415 & 0.317 & 0.350 & 0.406 & \underline{0.447} \\
       & R-L $\uparrow$ & 0.332 & \textbf{\underline{0.408}} & 0.398 & 0.358 & 0.270 & 0.305 & 0.356 & \underline{0.392} \\

\bottomrule
\end{tabular}}
\end{table}

\begin{table}[h]
\centering
\caption{Performance comparison between Qwen and DeepSeek-Qwen on LaMP tasks with RAG ($k=4$). The best results are highlighted in \textbf{bold}. }
\label{tab:lamp_results3}
\resizebox{\textwidth}{!}{
\begin{tabular}{l l c c c c c c c c}
\toprule
Task & Metric & Qwen1.5 & Qwen7 & Qwen14 & Qwen32 & \stackanchor{DeepSeek-}{Qwen1.5} & \stackanchor{DeepSeek-}{Qwen7} & \stackanchor{DeepSeek-}{Qwen14} & \stackanchor{DeepSeek-}{Qwen32} \\
\midrule
LaMP-1 & Acc $\uparrow$ & 0.595 & \textbf{\underline{0.765}} & 0.740 & 0.750 & 0.605 & 0.723 & \underline{0.742} & 0.733 \\
       & F1 $\uparrow$  & 0.506 & \textbf{\underline{0.764}} & 0.738 & 0.743 & 0.602 & 0.720 & \underline{0.741} & 0.729 \\
\midrule
LaMP-2N & Acc $\uparrow$ & 0.690 & 0.750 & 0.760 & \textbf{\underline{0.815}} & 0.567 & 0.717 & 0.763 & \underline{0.787} \\
       & F1 $\uparrow$  & 0.373 & 0.444 & 0.474 & \textbf{\underline{0.512}} & 0.263 & 0.421 & \underline{0.480} & 0.477 \\
\midrule
LaMP-2M & Acc $\uparrow$ & 0.320 & 0.410 & \underline{0.470} & 0.460 & 0.202 & 0.300 & 0.470 & \textbf{\underline{0.487}} \\
       & F1 $\uparrow$  & 0.231 & 0.256 & 0.363 & \underline{0.367} & 0.141 & 0.252 & 0.359 & \textbf{\underline{0.385}} \\
\midrule
LaMP-3 & MAE $\downarrow$ & 0.480 & 0.360 & \underline{0.240} & 0.245 & 0.715 & 0.332 & 0.212 & \textbf{\underline{0.200}} \\
       & RMSE $\downarrow$ & 0.760 & 0.450 & \underline{0.280} & 0.285 & 1.398 & 0.392 & 0.232 & \textbf{\underline{0.217}} \\
\midrule
LaMP-4 & R-1 $\uparrow$ & \textbf{\underline{0.174}} & 0.171 & 0.159 & 0.157 & 0.130 & 0.151 & \textbf{\underline{0.174}} & 0.170 \\
       & R-L $\uparrow$ & 0.145 & \underline{0.153} & 0.144 & 0.140 & 0.117 & 0.132 & \textbf{\underline{0.155}} & 0.150 \\
\midrule
LaMP-5 & R-1 $\uparrow$ & \textbf{\underline{0.469}} & 0.464 & 0.374 & 0.368 & 0.356 & 0.402 & \underline{0.449} & 0.433 \\
       & R-L $\uparrow$ & \textbf{\underline{0.404}} & 0.398 & 0.309 & 0.313 & 0.291 & 0.329 & \underline{0.382} & 0.364 \\
\midrule
LaMP-7 & R-1 $\uparrow$ & 0.407 & 0.450 & \underline{0.455} & 0.419 & 0.307 & 0.393 & 0.448 & \textbf{\underline{0.464}} \\
       & R-L $\uparrow$ & 0.357 & 0.392 & \underline{0.398} & 0.361 & 0.260 & 0.339 & 0.394 & \textbf{\underline{0.410}} \\

\bottomrule
\end{tabular}}
\end{table}

\begin{table}[h]
\centering
\caption{Average token length comparison of Qwen and DeepSeek-Qwen on LaMP Tasks with RAG ($k=1$).}
\label{tab:lamp_length_results2}
\resizebox{\textwidth}{!}{
\begin{tabular}{l c c c c c c c c}
\toprule
Task & Qwen1.5 & Qwen7 & Qwen14 & Qwen32 & \stackanchor{DeepSeek-}{Qwen1.5} & \stackanchor{DeepSeek-}{Qwen7} & \stackanchor{DeepSeek-}{Qwen14} & \stackanchor{DeepSeek-}{Qwen32} \\
\midrule
LaMP-1 & 3.0 & 3.0 & 3.0 & 4.2 & 537.8 & 730.9 & 449.5 & 351.6 \\
LaMP-2N & 2.1 & 1.8 & 3.3 & 1.8 & 300.3 & 283.4 & 244.6 & 258.1 \\
LaMP-2M & 3.4 & 2.4 & 5.1 & 2.7 & 363.9 & 460.7 & 339.7 & 384.7 \\
LaMP-3 & 1.0 & 1.0 & 1.0 & 1.0 & 398.8 & 348.3 & 295.5 & 314.6 \\
LaMP-4 & 16.9 & 38.3 & 26.8 & 42.5 & 448.0 & 446.7 & 412.8 & 386.9 \\
LaMP-5 & 19.9 & 22.8 & 35.5 & 49.2 & 477.1 & 381.4 & 410.8 & 376.1 \\
LaMP-7 & 24.8 & 24.6 & 25.5 & 27.0 & 394.9 & 402.8 & 432.0 & 377.6 \\

\midrule
Avg. & 10.2 & 13.4 & 14.3 & 18.3 & 417.3 & 436.3 & 369.3 & 349.9 \\
\bottomrule
\end{tabular}}
\end{table}

\begin{table}[h]
\centering
\caption{Average token length comparison of Qwen and DeepSeek-Qwen on LaMP Tasks with RAG ($k=4$).}
\label{tab:lamp_length_results_k4}
\resizebox{\textwidth}{!}{
\begin{tabular}{l c c c c c c c c}
\toprule
Task & Qwen1.5 & Qwen7 & Qwen14 & Qwen32 & \stackanchor{DeepSeek-}{Qwen1.5} & \stackanchor{DeepSeek-}{Qwen7} & \stackanchor{DeepSeek-}{Qwen14} & \stackanchor{DeepSeek-}{Qwen32} \\
\midrule
LaMP-1 & 3.0 & 3.0 & 3.0 & 3.0 & 650.4 & 597.1 & 359.8 & 374.6 \\
LaMP-2N & 1.9 & 1.9 & 1.8 & 1.8 & 313.6 & 326.6 & 254.0 & 263.1 \\
LaMP-2M & 2.7 & 2.1 & 2.8 & 2.1 & 446.6 & 498.1 & 353.8 & 374.6 \\
LaMP-3 & 1.0 & 1.0 & 1.0 & 1.0 & 437.1 & 397.6 & 343.7 & 370.1 \\
LaMP-4 & 17.5 & 21.9 & 30.4 & 31.2 & 505.5 & 470.4 & 435.2 & 393.1 \\
LaMP-5 & 18.9 & 21.0 & 40.9 & 40.7 & 573.0 & 432.1 & 512.6 & 444.0 \\
LaMP-7 & 25.4 & 24.7 & 25.6 & 27.5 & 554.7 & 463.5 & 408.8 & 363.2 \\

\midrule
Avg. & 10.1 & 10.8 & 15.1 & 15.3 & 497.3 & 455.1 & 381.1 & 369.0 \\
\bottomrule
\end{tabular}}
\end{table}

\section{Preliminary Evaluation}

\subsection{Evaluation Setup}
\label{sec:exp_setup}

In this section, we evaluate the performance of a general-purpose LLM and an LRM on the widely used LaMP dataset~\cite{salemi2024lamp}, including (1) Personalized Citation Identification (LaMP-1), (2) Personalized News Categorization (LaMP-2N), (3) Personalized Movie Tagging (LaMP-2M), (4) Personalized Product Rating (LaMP-3), (5) Personalized News Headline Generation (LaMP-4), (6) Personalized Scholarly Title Generation (LaMP-5), and (7) Personalized Tweet Paraphrasing (LaMP-7). We exclude the LaMP-6 task since this dataset is not publicly available. Similar to \cite{zhuanghydra}, we follow the user-based separation setting provided by the LaMP benchmark and randomly select 200 users for evaluation. Additional experimental details are provided in Appendix~\ref{sec:eval_detail}.

We select Llama-3.1-8B-Instruct~\cite{grattafiori2024llama} as the general-purpose LLM and DeepSeek-R1-Distill-Llama-8B~\cite{guo2025deepseek} as the LRM. For Llama-3.1-8B-Instruct, we set the temperature to 0 to ensure greedy decoding and consistent output. For DeepSeek-R1-Distill-Llama-8B, we set the temperature to 0.6, aligning with the official guidelines~\cite{deepseekr1}. 
We repeat the experiment three times and calculate the average.
We evaluate the models under four settings: non-personalized ($k=0$), random context ($k=1$), and retrieval-augmented generation (RAG)~\cite{salemi2024lamp} with $k=1$ and $k=4$ examples, where $k$ denotes the number of user-specific examples provided. We adopt BM25~\cite{robertson2009probabilistic} for all retrieval operations to ensure efficient and consistent comparisons.

Following LaMP \cite{salemi2024lamp}, we evaluate performance using accuracy and F1-score for classification tasks (LaMP-1, LaMP-2N, and LaMP-2M), and MAE and RMSE for regression tasks (LaMP-3). For text generation tasks (LaMP-4, LaMP-5, and LaMP-7), we adopt ROUGE-1 (R-1) and ROUGE-L (R-L) \cite{lin2004rouge}. Note that higher values indicate better performance for all metrics, except for RMSE and MAE, where lower values are better.

\subsection{Comparison between LLM and LRM}
\label{sec:comp}

We compare Llama-3.1-8B-Instruct (Llama3) and DeepSeek-R1-Distill-Llama-8B (DeepSeek-Llama3) under various settings.
An example of answers generated by Llama3 and DeepSeek-Llama3 is shown in Figure \ref{fig:appen}.
The results are presented in Table~\ref{tab:lamp_results1}. Surprisingly, we observe the general-purpose LLM (Llama3)  outperforms the LRM (DeepSeek-Llama) across most tasks in retrieval-intensive settings (RAG with $k=4$). For instance, in LaMP-1, Llama3 achieves an accuracy of 0.760 with RAG ($k=4$), compared to DeepSeek’s 0.712. However, in the RAG with $k=1$ setting, DeepSeek outperforms Llama3 in most tasks.
We hypothesize that general-purpose LLMs benefit more from in-context learning when provided with more relevant user-specific examples, leveraging these examples to adapt flexibly to diverse tasks. In contrast, LRMs may prioritize divergent or exploratory reasoning, potentially overlooking critical user profile details or failing to align outputs with expected formats.

To investigate this performance gap, our detailed analysis uncovers three critical limitations of LRMs in personalization tasks.


\begin{itemize}[left=0em]
    \item \textbf{Limited Divergent Thinking}: LRMs are optimized for convergent reasoning, excelling in tasks with well-defined solutions (e.g., solving equations). However, personalization often demands divergent thinking, which is the ability to explore multiple intermediate hypotheses and synthesize user-specific nuances. LRMs frequently fail to generate the creative, exploratory steps necessary to capture subtle user preferences, leading to suboptimal outputs.
    \item \textbf{Poor Response Format Alignment}: Personalization tasks often require outputs to adhere to specific formats. LRMs, trained primarily for correctness in reasoning, tend to produce inconsistent or unstructured responses, undermining their utility in practical settings.
    \item \textbf{Inefficient Utilization of Retrieved Knowledge}: While RAG frameworks enhance LLMs by providing external context, LRMs struggle to integrate retrieved information effectively. Their reasoning processes often prioritize internal logic over contextual relevance, resulting in outputs that fail to reflect user-specific insights embedded in the retrieved data.
\end{itemize}

We quantify each of these limitations in Section~\ref{sec:quant_limitations}.

\textbf{\textsc{Takeaway I:}} 
General-purpose LLMs surpass LRMs on more tasks under retrieval-intensive setting, likely due to stronger in-context learning ability.

\subsection{Comparison between Model Scaling}
\label{subsec:scaling}

To investigate the impact of model size on personalization performance, we evaluate Qwen2.5-Instruct series models~\cite{yang2024qwen2} (with 1.5B, 7B, 14B, 32B parameters) and their DeepSeek-Distill-Qwen counterparts \cite{deepseekr1}, which are optimized for reasoning, on LaMP tasks under RAG settings ($k=1$ and $k=4$). 
The Qwen2.5-Instruct-1.5B model is abbreviated as Qwen 1.5, and similarly for other sizes.
Results are reported in Tables~\ref{tab:lamp_results2} and~\ref{tab:lamp_results3}, respectively.

We could observe neither Qwen models nor DeepSeek-Distill-Qwen models consistently dominate all tasks.
Qwen models often outperform DeepSeek-Distill-Qwen models in generation tasks, particularly in the few-shot setting. Conversely, DeepSeek-Qwen models tend to excel in classification and regression tasks, especially with increased context.
For the DeepSeek-Qwen models, larger models generally outperform smaller ones across most tasks. This verifies that the enhanced reasoning ability of larger models positively impacts the personalization tasks.
Moreover, increasing context from \( k=1 \) to \( k=4 \) generally improves performance for both base and DeepSeek-Qwen models across most tasks, with more pronounced gains in classification and regression tasks.

\textbf{\textsc{Takeaway II:}} Larger LRMs 
generally outperform smaller ones due to enhanced reasoning capability.


\subsection{Impact of Generated Token Length}

We analyze the trends in average token length by comparing the Llama, Qwen, and DeepSeek series models under different settings. The results are presented in Table \ref{tab:lamp_length_results}, \ref{tab:lamp_length_results2}, and \ref{tab:lamp_length_results_k4}.

Our findings indicate that larger reasoning models generate significantly longer token lengths compared to their base counterparts (Qwen and Llama) across all tasks and settings. Additionally, token lengths vary considerably across different tasks, with generation tasks typically yielding longer outputs than classification or regression tasks for both the base models and DeepSeek models.
In Tables~\ref{tab:lamp_length_results2} and \ref{tab:lamp_length_results_k4}, we observe that larger language models tend to produce slightly longer token lengths than smaller ones, while the opposite trend is noted for the larger reasoning models (LRM). An increase in context size from $k=1$ to $k=4$ generally results in slightly longer token lengths for DeepSeek models, particularly for smaller LRMs.

\textbf{\textsc{Takeaway III:}} 
Larger general LLMs produce longer responses, while
larger LRMs tend to generate shorter yet more effective reasoning paths.


\subsection{Quantifying the Three Limitations}
\label{sec:quant_limitations}

Section~\ref{sec:comp} identified three limitations of LRMs in personalization
tasks. Here we make each of them measurable, using the generations already
produced for Tables~\ref{tab:lamp_results1}--\ref{tab:lamp_results3} together
with a set of controlled reruns. Unless stated otherwise, all numbers use the
same 200 users, BM25 retrieval and $k=4$.

\textbf{Poor Response Format Alignment.}
We define a generation as \emph{format-violating} if, after removing the
reasoning trace, the remaining text is not exactly a member of the task's label
set (\texttt{[1]}/\texttt{[2]} for LaMP-1, one of the 15 category names for
LaMP-2N/2M, a single digit in $\{1,\dots,5\}$ for LaMP-3), so that a downstream
system would need a heuristic parser to recover the answer. Results are shown in
Table~\ref{tab:format_violation}. The general-purpose LLM almost never violates
the format ($1.3\%$ on average), whereas the LRM does so for a third of all
queries ($33.8\%$), peaking at $55.5\%$ on LaMP-2M. The hierarchical reasoning
thought template together with the reasoning process intervention roughly halves
the violation rate ($17.1\%$), which quantifies the format-alignment benefit of
the structured pipeline. The last two rows show that the effect
also shrinks with backbone scale, mirroring \textsc{Takeaway II}.

\begin{table}[t]
\centering
\caption{Format-violation rate (\%, lower is better): fraction of generations
whose final answer is not directly a valid label of the task. Computed on the
generations behind Tables~\ref{tab:lamp_results1}--\ref{tab:lamp_results3}.}
\label{tab:format_violation}
\resizebox{0.85\textwidth}{!}{
\begin{tabular}{l c c c c c}
\toprule
Model & LaMP-1 & LaMP-2N & LaMP-2M & LaMP-3 & Avg. \\
\midrule
Llama3 (RAG, $k=1$)          &  0.0 &  3.5 &  5.0 & 0.0 &  2.1 \\
Llama3 (RAG, $k=4$)          &  0.0 &  2.0 &  3.0 & 0.0 &  1.3 \\
\midrule
DeepSeek-Llama3 (RAG, $k=1$) & 42.0 & 23.5 & 57.0 & 2.5 & 31.3 \\
DeepSeek-Llama3 (RAG, $k=4$) & 51.0 & 25.0 & 55.5 & 3.5 & 33.8 \\
\rowcolor{cyan!20} DeepSeek-Llama3 (+ HRT + RPI, $k=4$) & \textbf{21.5} & \textbf{12.5} & \textbf{30.0} & 4.5 & \textbf{17.1} \\
\midrule
DeepSeek-Qwen7 (RAG, $k=4$)   & 22.0 & 1.5 & 0.5 & 1.0 & 6.3 \\
DeepSeek-Qwen14 (RAG, $k=4$)  & 11.5 & 1.0 & 0.5 & 2.5 & 3.9 \\
\bottomrule
\end{tabular}}
\end{table}

\textbf{Undirected Exploration.}
The limitation identified in Section~\ref{sec:comp} is not that LRMs fail to
explore, but that their exploration is not directed at the user: the trace
repeatedly interrupts itself -- proposing a hypothesis, abandoning it and
restarting -- while rarely reasoning about the user profile at all (see the last
paragraph of this section). We measure this undirected exploration by counting
backtracking markers (\textit{wait}, \textit{hmm}, \textit{alternatively},
\textit{however}, \textit{actually}, \textit{but}, \textit{maybe},
\textit{let me reconsider}) and normalising by trace length.
Table~\ref{tab:divergence} reports the rate per 1k words. The hierarchical
reasoning thought template reduces backtracking on all seven tasks, by $31\%$ on
average, confirming that the template redirects the reasoning budget rather than
merely shortening the output: the same trace spends less of itself on
self-interruption and more of it on the prescribed subtasks. Adding RPI raises the rate slightly,
by design: the intervention re-opens the chain exactly once, and the resulting
rate remains far below plain RAG prompting. The bottom block shows that a larger
LRM backtracks less than a smaller one under identical conditions, giving a
mechanism for \textsc{Takeaway III}.

\begin{table}[t]
\centering
\caption{Backtracking markers per 1k words in the reasoning trace (lower means
less undirected exploration). Top block: DeepSeek-Llama3 under identical
decoding conditions. Bottom block: backbone scale, measured on the runs behind
Table~\ref{tab:lamp_results3}.}
\label{tab:divergence}
\resizebox{\textwidth}{!}{
\begin{tabular}{l c c c c c c c c}
\toprule
Configuration & LaMP-1 & LaMP-2N & LaMP-2M & LaMP-3 & LaMP-4 & LaMP-5 & LaMP-7 & Avg. \\
\midrule
DeepSeek-Llama3 (RAG)              & 15.7 & 10.3 & 19.0 & 15.0 & 17.0 & 18.4 & 15.6 & 15.9 \\
DeepSeek-Llama3 (+ HRT)            & \textbf{8.5} & \textbf{7.3} & \textbf{12.9} & \textbf{13.2} & \textbf{13.4} & \textbf{11.2} & \textbf{10.5} & \textbf{11.0} \\
\rowcolor{cyan!20} DeepSeek-Llama3 (+ HRT + RPI)& 11.8 & 7.7 & 15.1 & 15.2 & 14.7 & 12.8 & 11.4 & 12.7 \\
\midrule
DeepSeek-Qwen7 (RAG)               & 16.8 & 11.3 & 24.6 & 18.7 & 25.8 & 14.8 & 25.8 & 19.7 \\
DeepSeek-Qwen14 (RAG)              & 5.0 & 10.4 & 15.4 & 14.3 & 15.4 & 12.9 & 17.3 & 13.0 \\
\bottomrule
\end{tabular}}
\end{table}

\textbf{Output Inconsistency.}
The same exploratory tendency also makes the LRM unstable across runs. We repeat
each setting three times (as in Section~\ref{sec:exp_setup}) and report the mean
pairwise agreement between runs, i.e. the fraction of users for which two runs
emit exactly the same answer (Table~\ref{tab:self_consistency}). The LRM is
markedly less consistent than the general-purpose LLM on every task, and on the
three generation tasks it essentially never reproduces its own output
($0.0$--$1.5\%$ versus $11.2$--$20.7\%$). This is precisely the failure mode the
self-referencing module of Section~4.3 is designed to absorb,
and it explains why majority voting alone is insufficient for generation tasks.

\begin{table}[t]
\centering
\caption{Mean pairwise agreement between three repeated runs (\%, higher means
more self-consistent), $k=4$. Llama3 uses greedy decoding and DeepSeek-Llama3
uses $T=0.6$ following the official guidelines; the residual variation of
Llama3 comes from non-deterministic batched inference.}
\label{tab:self_consistency}
\resizebox{\textwidth}{!}{
\begin{tabular}{l c c c c c c c}
\toprule
Model & LaMP-1 & LaMP-2N & LaMP-2M & LaMP-3 & LaMP-4 & LaMP-5 & LaMP-7 \\
\midrule
Llama3          & 88.0 & 91.3 & 76.7 & 84.3 & 15.8 & 20.7 & 11.2 \\
DeepSeek-Llama3 & 74.7 & 81.5 & 59.0 & 73.2 & \phantom{0}1.5 & \phantom{0}0.0 & \phantom{0}0.7 \\
\bottomrule
\end{tabular}}
\end{table}

\textbf{Inefficient Utilization of Retrieved Knowledge.}
Finally, we apply the checklist of Section~4.2 to the reasoning
traces produced by plain RAG prompting: a trace is flagged when it never
mentions the user profile or the user intent. Across all seven tasks,
$97.4\%$ of the traces at $k=4$ ($96.9\%$ at $k=1$) are flagged, i.e. the LRM
reaches its answer without ever explicitly considering who the user is, even
though $k$ profile entries are present in its context. In our controlled reruns the same statistic falls from $95.9\%$ for RAG
prompting to $92.1\%$ once the hierarchical template is added and to $79.3\%$
once the reasoning process intervention is applied on top of it. This is the
most direct evidence that the retrieved user history is under-used, and it
motivates the reasoning process intervention introduced in
Section~4.2.

\textbf{\textsc{Takeaway IV:}}
The three limitations are measurable: at $k=4$ the LRM violates the required
output format on $33.8\%$ of queries, backtracks $15.9$ times per 1k words of
reasoning, reproduces its own answer on generation tasks in under $2\%$ of cases,
and never explicitly reasons about the user in $97.4\%$ of its traces.

\section{Methodology}

\begin{figure}[t]
\centering
\includegraphics[width=\textwidth]{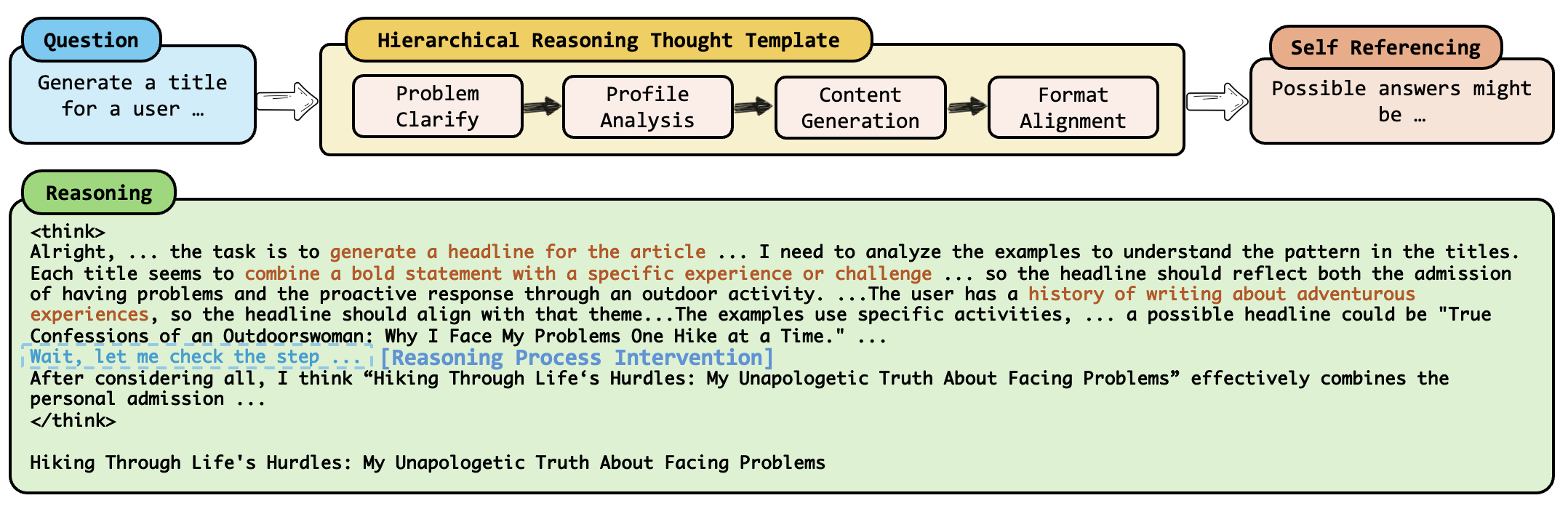}
    \caption{Pipeline of the proposed \model. For a given query, the model integrates a reasoning thought template to guide structured reasoning, employs self-referencing to ensure consistency, and incorporates reasoning process intervention to maintain alignment with task requirements.}
\label{fig:model}
\end{figure}

We propose \textbf{Reinforced Reasoning for Personalization (R2P)}, a novel framework designed to enhance the performance of LRMs in personalization tasks. R2P combines \textit{hierarchical reasoning thought template}, which decomposes complex personalization tasks into structured subtasks, with a \textit{reasoning process intervention} that reinforces the reasoning process to ensure alignment with the desired reasoning path. 
The self referencing module is introduced to ensure consistency.
The pipeline of the proposed \model is presented in Figure \ref{fig:model}.

\subsection{Hierarchical Reasoning Thought Template}

As analyzed in Section \ref{sec:comp}, LRMs often struggle with personalization tasks due to their tendency toward divergent thinking and inefficient use of RAG information. Without a structured framework, LRMs may produce outputs that lack focus, deviate from user-specific requirements, or underutilize critical contextual data, leading to suboptimal personalization performance.

To solve this, we propose a hierarchical reasoning thought template, which offers a structured approach to guide LRMs in personalization tasks. By providing a clear framework, it ensures critical RAG information is utilized effectively, enhancing the focus and consistency of outputs. The template, which can be handcrafted or generated by an LLM, streamlines the reasoning process into a concise pipeline: first, analyzing the input to identify user-specific requirements; then, synthesizing a user profile from preferences, interactions, or contextual cues; and finally, generating a tailored, contextually aligned response. This approach curbs divergent thinking, ensures adherence to user constraints, and boosts personalization performance by fully leveraging RAG information. A prompt example is shown in Appendix \ref{sec:prompt}.

The HRT addresses the limitations of LRMs by enforcing a structured pipeline that curbs excessive divergent thinking and ensures alignment with task specifications. By explicitly prioritizing retrieved context, the HRT also improves the utilization of RAG data, overcoming the inefficiencies observed in baseline LRMs.

\subsection{Reasoning Process Intervention}

Even with a structured template, LRMs may deviate from the intended reasoning path due to their tendency to prioritize internal logic over the provided template. To address this, we introduce a Reasoning Process Intervention (RPI) mechanism that dynamically monitors and corrects the reasoning process. RPI operates as a feedback loop, ensuring that all subtasks of the HRT are executed correctly and that the output aligns with personalization goals.

The RPI mechanism functions as follows: after the LRM completes the reasoning chain, we evaluate the output against a checklist derived from the HRT. This checklist verifies whether each subtask (e.g., query decomposition, profile synthesis) has been adequately addressed and whether the response adheres to the required format and user preferences.
If deviations are detected (e.g., missing user profile analysis), RPI intervenes at the end of the reasoning process and injects an instruction such as, “Wait, let me analyze the user profile.” The LRM is then instructed to revise its reasoning chain, starting from the flagged subtask, while preserving valid components of the previous output. In this way, RPI enhances model performance by preventing LRMs from producing misaligned outputs.

\subsection{Self Referencing Module}

Personalization tasks often require consistency across multiple outputs to ensure coherence and reliability. However, LRMs may produce varied responses for similar inputs due to their exploratory reasoning tendencies. For classification tasks, majority voting \cite{wangself} can be effectively employed. However, for generative tasks, majority voting cannot be directly applied.

To address this challenge, we propose a Self-Referencing Module (SRM) to refine outputs.
The SRM operates as follows: for a given query, the LRM generates multiple candidate responses (e.g., $n$ outputs) using the HRT and RPI. Each response incorporates slight variations in the reasoning chain to explore different personalization angles. These responses are then integrated into the prompt sent to the LRM for generating a new output. The prompt is structured as: “Possible answers might be: [Response 1], ..., [Response $n$].”
The LRM is provided with all candidate responses and instructed to synthesize a final output that balances the strengths of each while adhering to the HRT. We employ a collective self-referencing approach, which allows for parallelization. Alternatively, an iterative self-referencing method could be used, where for each round, the LRM generates answers for the next round of reference.
In this way, the SRM ensures that the final output is both personalized and coherent, effectively addressing the divergent thinking limitations of LRMs.

\section{Experiment}

\subsection{Experiment Setup}

We adopt the same experimental setup as described in Section~\ref{sec:exp_setup}.
We compare our approach with several baseline methods, including retrieval-augmented prompting (RAG)~\cite{salemi2024lamp}, profile-augmented prompting (PAG)~\cite{richardson2023integrating}, as well as self-verification \cite{weng2023large}.
Following \cite{zhuanghydra}, we exclude LaMP-7 due to its inconsistent user history and query format.
The reference number is set to $n=2$.
By default, we employ the RAG style across all experiments.

\begin{table}[t]
\centering
\caption{Main experiment results on the LaMP benchmark. The best scores for each
task are highlighted in \textbf{boldface}. * denotes for $p<0.05$ based on a paired t-test.}
\label{tab:lamp_main_results}
\resizebox{\textwidth}{!}{
\begin{tabular}{c l *{12}{c}}
\toprule
\multirow{2}{*}{Setting} & Task ($\rightarrow$) & \multicolumn{2}{c}{LaMP-1} & \multicolumn{2}{c}{LaMP-2N} & \multicolumn{2}{c}{LaMP-2M} & \multicolumn{2}{c}{LaMP-3} & \multicolumn{2}{c}{LaMP-4} & \multicolumn{2}{c}{LaMP-5} \\
\cmidrule(lr){3-4} \cmidrule(lr){5-6} \cmidrule(lr){7-8} \cmidrule(lr){9-10} \cmidrule(lr){11-12} \cmidrule(lr){13-14}
 & Method ($\downarrow$) & Acc $\uparrow$ & F1 $\uparrow$ & Acc $\uparrow$ & F1 $\uparrow$ & Acc $\uparrow$ & F1 $\uparrow$ & MAE $\downarrow$ & RMSE $\downarrow$ & R-1 $\uparrow$ & R-L $\uparrow$ & R-1 $\uparrow$ & R-L $\uparrow$ \\
\midrule
\multirow{4}{*}{$k=1$} & RAG & 0.650 & 0.649 & 0.698 & 0.421 & \textbf{0.332} & \textbf{0.257} & 0.383 & 0.678 & 0.152 & 0.133 & 0.406 & 0.341 \\
 & PAG & 0.648 & 0.648 & \textbf{0.715} & 0.398 & 0.298 & 0.227 & 0.395 & 0.717 & 0.143 & 0.125 & 0.300 & 0.253 \\
 & Self-Verification & 0.613 & 0.611 & 0.713 & \textbf{0.425} & 0.307 & 0.209 & 0.432 & 0.746 & 0.117 & 0.103 & 0.249 & 0.210 \\
\rowcolor{cyan!20} \cellcolor{white} & \model & \textbf{0.673*} & \textbf{0.673} & 0.697 & 0.394 & 0.282 & 0.200 & \textbf{0.345} & \textbf{0.637} & \textbf{0.155} & \textbf{0.135} & \textbf{0.412} & \textbf{0.342} \\
 \midrule
\multirow{4}{*}{$k=4$} & RAG & 0.712 & 0.711 & 0.697 & 0.392 & 0.403 & 0.287 & 0.325 & 0.664 & 0.166 & 0.148 & 0.413 & 0.344 \\
 & PAG & 0.710 & 0.710 & \textbf{0.722} & 0.401 & \textbf{0.437} & 0.318 & 0.280 & 0.619 & 0.164 & 0.147 & 0.344 & 0.289 \\
 & Self-Verification & 0.695 & 0.694 & 0.713 & 0.406 & 0.395 & 0.284 & 0.322 & 0.633 & 0.125 & 0.109 & 0.266 & 0.225 \\
\rowcolor{cyan!20} \cellcolor{white} &  \model & \textbf{0.743*} & \textbf{0.742*} & 0.720 & \textbf{0.417} & 0.427 & \textbf{0.319} & \textbf{0.257*} & \textbf{0.559*} & \textbf{0.174*} & \textbf{0.154} & \textbf{0.429} & \textbf{0.356} \\

\bottomrule
\end{tabular}}
\vspace{-0.05in}
\end{table}

\begin{figure}[t]
\centering
\includegraphics[width=\textwidth]{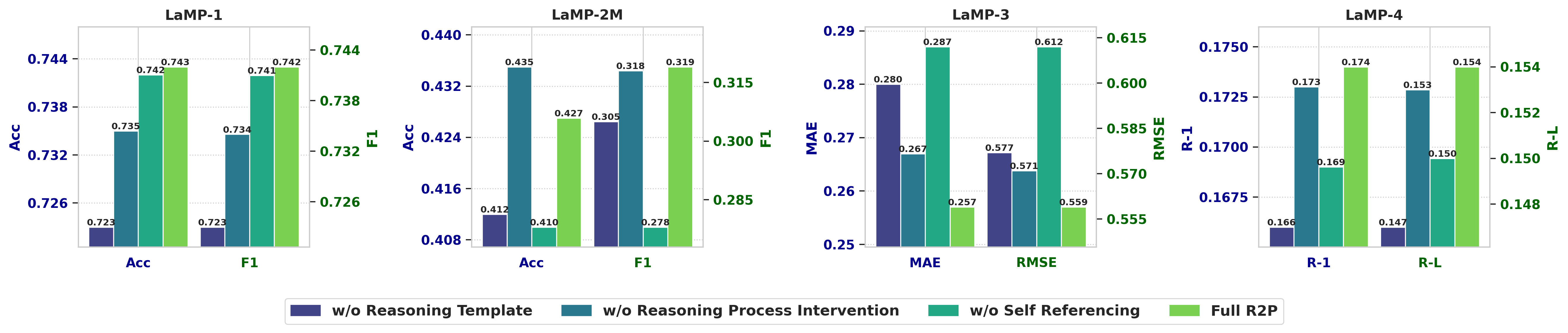}
    \caption{Ablation study of \model across four LaMP tasks.}
\label{fig:abaltion}
\vspace{-0.2in}
\end{figure}

\subsection{Main Results}
We evaluate \model against baseline methods on few-shot ($k=1$) and multi-shot ($k=4$) settings. 
The main results are reported in Table~\ref{tab:lamp_main_results}.
We can observe proposed \model method achieves the best performance on most of the tasks.
With more context, \model consistently outperforms baselines across more tasks and achieves greater performance improvements.
Notably, \model excels in regression and generation tasks, where it consistently achieves the lowest errors and highest ROUGE scores. For classification tasks, \model dominates LaMP-1 and shows competitive performance on LaMP-2N and LaMP-2M, particularly in the multi-shot setting. The baselines, while occasionally competitive, generally lag behind \model, especially as context increases. These results highlight \model's effectiveness in personalization tasks, leveraging context to improve performance across diverse evaluation metrics.

\subsection{Ablation Study}

We conduct an ablation study using the full \model alongside three variants: without reasoning template, without reasoning process intervention, and without self-referencing. The results are presented in Figure \ref{fig:abaltion}. We observe that the full \model achieves the best results in most cases, which verifies the effectiveness and importance of each module. Additionally, we note that removing the reasoning template leads to a significant performance drop, validating the importance of the guided LRM in generating structured output.
Moreover, removing the reasoning process intervention can degrade model performance. This may be because the model does not align well with the instructions and needs intervention to better focus on the given template.

\begin{figure}[t]
\centering
\includegraphics[width=\textwidth]{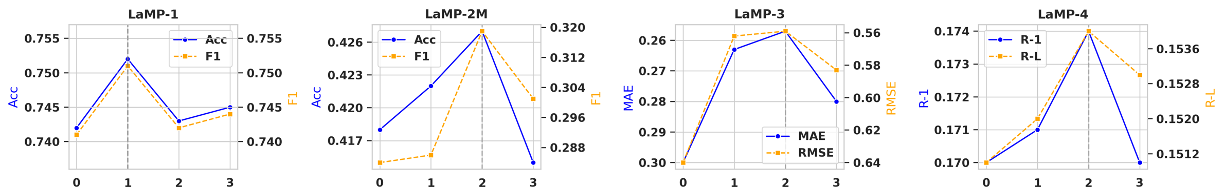}
    \caption{Performance under different $n$ across four LaMP tasks.}
\label{fig:study_n}
\end{figure}

\begin{figure}[t]
\centering
\includegraphics[width=\textwidth]{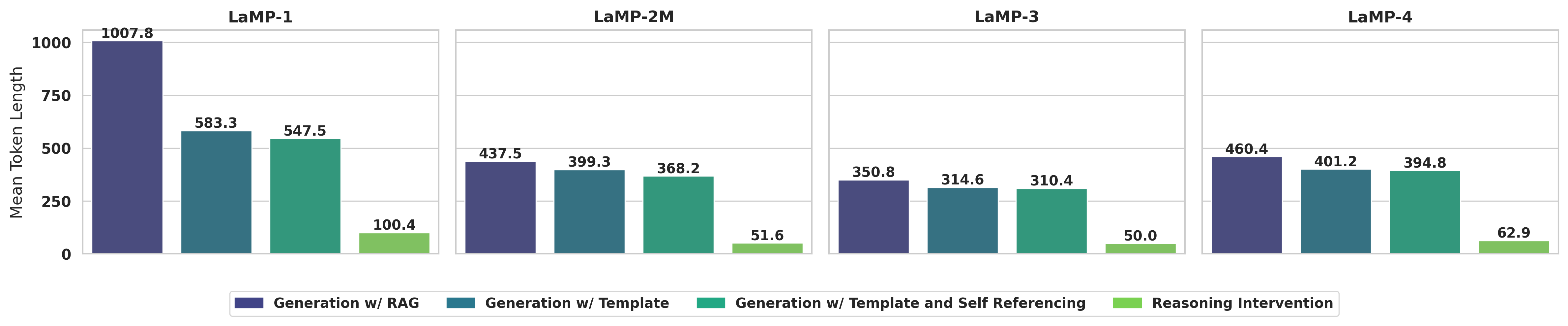}
    \caption{Mean token lengths of generated outputs across four LaMP tasks under different configurations.}
\label{fig:token_length}
\end{figure}

\subsection{Further Analysis}

\textbf{Analysis of Hyperparameter \( n \) on Model Performance.}
We analyze the impact of the hyperparameter \( n \), ranging from 0 to 3, on our proposed method across selected tasks. The result is shown in Figure~\ref{fig:study_n}.
We could observe that
the hyperparameter \( n=2 \) emerges as the most effective setting, achieving the best performance in three out of four tasks and competitive results in LaMP-1. This suggests that \( n=2 \) provides an optimal configuration for balancing model complexity and task-specific performance across classification, regression, and generation tasks. Notably, \( n=1 \) performs best for LaMP-1, indicating task-specific sensitivity to \( n \), particularly for simpler classification tasks. Higher values (\( n=3 \)) generally lead to slight performance degradation, while \( n=0 \) consistently underperforms, underscoring the importance of self referencing.

\textbf{Analysis of Computation Cost of \model.}
We analyze the mean token lengths of outputs generated by LRM across four LaMP tasks under different configurations, including standard RAG, generation with hierarchical reasoning thought template, generation with template and self-referencing, and generation with reasoning process intervention. Results are presented in Figure~\ref{fig:token_length}.

Our findings reveal that incorporating a reasoning template reduces the mean token length compared to original RAG. Adding self-referencing further decreases token length, and the additional computational cost of reasoning intervention remains low. This observation is interesting, as standard RAG, without structured guidance, often leads to inefficient thinking, resulting in longer outputs and poorer performance. In contrast, generation with reasoning template configurations produce shorter, more focused outputs with improved performance.
These results suggest that generating fewer tokens correlates with better outcomes, challenging the assumption that longer outputs are inherently better.
However, as our approach involves multiple steps, the overall computational cost may exceed that of standard RAG, despite achieving superior performance. In resource constrained scenarios, using the reasoning template configuration alone strikes an effective balance between efficiency and performance, offering a practical solution for personalization tasks.

\subsection{Generalization to Other Reasoning Backbones}
\label{sec:backbone}

The results in Table~\ref{tab:lamp_main_results} are obtained with
DeepSeek-R1-Distill-Llama-8B. To test whether the gains of \model are tied to
that particular backbone, we rerun the two prompting components of \model --
the hierarchical reasoning thought template (HRT) and the reasoning process
intervention (RPI) -- on DeepSeek-R1-Distill-Qwen-7B and
DeepSeek-R1-Distill-Qwen-14B under the same protocol (BM25, $k=4$, three
repetitions, $T=0.6$); the RAG rows are the corresponding results of
Table~\ref{tab:lamp_results3}. The self-referencing module is not applied here,
so this study isolates the contribution of the reasoning-side components.
Results are reported in Table~\ref{tab:backbone}.

On the 14B backbone HRT+RPI improves six of the seven tasks, most visibly on
LaMP-3, where MAE drops from $0.235$ to $0.193$ ($-17.9\%$), and on the three
generation tasks (R-1 $+0.018$, $+0.012$ and $+0.016$ respectively). On the 7B
backbone the picture is mixed: LaMP-1 improves substantially ($0.675 \rightarrow
0.747$ accuracy) and LaMP-5 gains $+0.032$ R-1, but LaMP-2N and LaMP-3
degrade. Together with Table~\ref{tab:lamp_main_results}, this indicates that
the benefit of a structured reasoning template grows with the reasoning ability
of the underlying model: a stronger LRM follows the template faithfully and
gains across the board, whereas a weaker one applies it inconsistently. This is
consistent with \textsc{Takeaway II} and suggests that the structured reasoning
of \model becomes more, not less, useful as reasoning models improve.

\begin{table}[t]
\centering
\caption{The reasoning-side components of \model on other reasoning backbones
(BM25, $k=4$, averaged over three runs). RAG rows are taken from
Table~\ref{tab:lamp_results3}; the self-referencing module is not applied in
this study. LaMP-7 is reported here for completeness, although it is excluded
from Table~\ref{tab:lamp_main_results} following \cite{zhuanghydra}. Best
result per backbone and task in \textbf{bold}.}
\label{tab:backbone}
\resizebox{\textwidth}{!}{
\begin{tabular}{c l *{14}{c}}
\toprule
\multirow{2}{*}{Backbone} & Task ($\rightarrow$) & \multicolumn{2}{c}{LaMP-1} & \multicolumn{2}{c}{LaMP-2N} & \multicolumn{2}{c}{LaMP-2M} & \multicolumn{2}{c}{LaMP-3} & \multicolumn{2}{c}{LaMP-4} & \multicolumn{2}{c}{LaMP-5} & \multicolumn{2}{c}{LaMP-7} \\
\cmidrule(lr){3-4} \cmidrule(lr){5-6} \cmidrule(lr){7-8} \cmidrule(lr){9-10} \cmidrule(lr){11-12} \cmidrule(lr){13-14} \cmidrule(lr){15-16}
 & Method ($\downarrow$) & Acc $\uparrow$ & F1 $\uparrow$ & Acc $\uparrow$ & F1 $\uparrow$ & Acc $\uparrow$ & F1 $\uparrow$ & MAE $\downarrow$ & RMSE $\downarrow$ & R-1 $\uparrow$ & R-L $\uparrow$ & R-1 $\uparrow$ & R-L $\uparrow$ & R-1 $\uparrow$ & R-L $\uparrow$ \\
\midrule
\multirow{3}{*}{\stackanchor{DeepSeek-}{Qwen7}}
 & RAG        & 0.675 & 0.667 & \textbf{0.690} & \textbf{0.425} & \textbf{0.300} & \textbf{0.255} & \textbf{0.345} & \textbf{0.689} & 0.147 & 0.129 & 0.382 & 0.320 & 0.372 & 0.323 \\
 & \, + HRT & 0.738 & 0.734 & 0.660 & 0.381 & 0.293 & 0.228 & 0.410 & 0.764 & 0.156 & 0.137 & \textbf{0.419} & \textbf{0.347} & \textbf{0.390} & \textbf{0.340} \\
\rowcolor{cyan!20} \cellcolor{white} & \, + HRT + RPI & \textbf{0.747} & \textbf{0.742} & 0.668 & 0.388 & 0.293 & 0.227 & 0.383 & 0.741 & \textbf{0.157} & \textbf{0.138} & 0.414 & 0.344 & 0.385 & 0.337 \\
\midrule
\multirow{3}{*}{\stackanchor{DeepSeek-}{Qwen14}}
 & RAG        & 0.765 & 0.763 & 0.765 & \textbf{0.481} & 0.470 & 0.358 & 0.235 & 0.579 & 0.156 & 0.136 & 0.459 & 0.387 & 0.443 & 0.391 \\
 & \, + HRT & \textbf{0.767} & \textbf{0.765} & 0.773 & 0.475 & \textbf{0.475} & \textbf{0.367} & \textbf{0.193} & \textbf{0.450} & \textbf{0.174} & \textbf{0.159} & \textbf{0.471} & \textbf{0.401} & 0.456 & 0.402 \\
\rowcolor{cyan!20} \cellcolor{white} & \, + HRT + RPI & 0.765 & 0.763 & \textbf{0.778} & 0.479 & 0.465 & 0.356 & 0.198 & 0.471 & \textbf{0.174} & 0.157 & 0.470 & 0.400 & \textbf{0.459} & \textbf{0.406} \\
\bottomrule
\end{tabular}}
\end{table}

\section{Conclusion}
In this work, we conduct the first systematic evaluation of large reasoning models (LRMs) for personalization tasks, revealing their unexpected underperformance compared to general-purpose LLMs, particularly in retrieval-intensive scenarios. Through detailed analysis, we identify three core limitations: limited divergent thinking, poor response format alignment, and inefficient use of retrieved knowledge, which hinder LRMs in personalization contexts. 
To address these limitations, we introduce Reinforced Reasoning for Personalization (\model), a novel training-free framework that leverages a hierarchical reasoning thought template, reasoning process intervention, and self referencing mechanism to guide LRMs toward generating structured outputs. Through extensive experiments, \model demonstrates superior performance compared with baseline methods.

\bibliographystyle{abbrv}
\bibliography{refs}

@inproceedings{salemi2024lamp,
  title={LaMP: When Large Language Models Meet Personalization},
  author={Salemi, Alireza and Mysore, Sheshera and Bendersky, Michael and Zamani, Hamed},
  booktitle={Proceedings of the 62nd Annual Meeting of the Association for Computational Linguistics (Volume 1: Long Papers)},
  pages={7370--7392},
  year={2024}
}

@article{grattafiori2024llama,
  title={The llama 3 herd of models},
  author={Grattafiori, Aaron and Dubey, Abhimanyu and Jauhri, Abhinav and Pandey, Abhinav and Kadian, Abhishek and Al-Dahle, Ahmad and Letman, Aiesha and Mathur, Akhil and Schelten, Alan and Vaughan, Alex and others},
  journal={arXiv preprint arXiv:2407.21783},
  year={2024}
}

@article{guo2025deepseek,
  title={Deepseek-r1: Incentivizing reasoning capability in llms via reinforcement learning},
  author={Guo, Daya and Yang, Dejian and Zhang, Haowei and Song, Junxiao and Zhang, Ruoyu and Xu, Runxin and Zhu, Qihao and Ma, Shirong and Wang, Peiyi and Bi, Xiao and others},
  journal={arXiv preprint arXiv:2501.12948},
  year={2025}
}

@inproceedings{zhuanghydra,
  title={HYDRA: Model Factorization Framework for Black-Box LLM Personalization},
  author={Zhuang, Yuchen and Sun, Haotian and Yu, Yue and Qiang, Rushi and Wang, Qifan and Zhang, Chao and Dai, Bo},
  booktitle={The Thirty-eighth Annual Conference on Neural Information Processing Systems}
}

@misc{deepseekr1,
  author = {{DeepSeek}},
  title = {DeepSeek-R1},
  year = {2025},
  howpublished = {\url{https://huggingface.co/deepseek-ai/DeepSeek-R1}}
}

@article{robertson2009probabilistic,
  title={The probabilistic relevance framework: BM25 and beyond},
  author={Robertson, Stephen and Zaragoza, Hugo and others},
  journal={Foundations and Trends{\textregistered} in Information Retrieval},
  volume={3},
  number={4},
  pages={333--389},
  year={2009},
  publisher={Now Publishers, Inc.}
}

@inproceedings{lin2004rouge,
  title={Rouge: A package for automatic evaluation of summaries},
  author={Lin, Chin-Yew},
  booktitle={Text summarization branches out},
  pages={74--81},
  year={2004}
}

@article{jaech2024openai,
  title={Openai o1 system card},
  author={Jaech, Aaron and Kalai, Adam and Lerer, Adam and Richardson, Adam and El-Kishky, Ahmed and Low, Aiden and Helyar, Alec and Madry, Aleksander and Beutel, Alex and Carney, Alex and others},
  journal={arXiv preprint arXiv:2412.16720},
  year={2024}
}

@article{wei2022chain,
  title={Chain-of-thought prompting elicits reasoning in large language models},
  author={Wei, Jason and Wang, Xuezhi and Schuurmans, Dale and Bosma, Maarten and Xia, Fei and Chi, Ed and Le, Quoc V and Zhou, Denny and others},
  journal={Advances in neural information processing systems},
  volume={35},
  pages={24824--24837},
  year={2022}
}

@article{yao2023tree,
  title={Tree of thoughts: Deliberate problem solving with large language models},
  author={Yao, Shunyu and Yu, Dian and Zhao, Jeffrey and Shafran, Izhak and Griffiths, Tom and Cao, Yuan and Narasimhan, Karthik},
  journal={Advances in neural information processing systems},
  volume={36},
  pages={11809--11822},
  year={2023}
}

@inproceedings{huang2023towards,
  title={Towards Reasoning in Large Language Models: A Survey},
  author={Huang, Jie and Chang, Kevin Chen-Chuan},
  booktitle={Findings of the Association for Computational Linguistics: ACL 2023},
  pages={1049--1065},
  year={2023}
}

@article{xu2025towards,
  title={Towards Large Reasoning Models: A Survey of Reinforced Reasoning with Large Language Models},
  author={Xu, Fengli and Hao, Qianyue and Zong, Zefang and Wang, Jingwei and Zhang, Yunke and Wang, Jingyi and Lan, Xiaochong and Gong, Jiahui and Ouyang, Tianjian and Meng, Fanjin and others},
  journal={arXiv preprint arXiv:2501.09686},
  year={2025}
}

@article{chen2024large,
  title={When large language models meet personalization: Perspectives of challenges and opportunities},
  author={Chen, Jin and Liu, Zheng and Huang, Xu and Wu, Chenwang and Liu, Qi and Jiang, Gangwei and Pu, Yuanhao and Lei, Yuxuan and Chen, Xiaolong and Wang, Xingmei and others},
  journal={World Wide Web},
  volume={27},
  number={4},
  pages={42},
  year={2024},
  publisher={Springer}
}

@article{zhang2024personalization,
  title={Personalization of large language models: A survey},
  author={Zhang, Zhehao and Rossi, Ryan A and Kveton, Branislav and Shao, Yijia and Yang, Diyi and Zamani, Hamed and Dernoncourt, Franck and Barrow, Joe and Yu, Tong and Kim, Sungchul and others},
  journal={arXiv preprint arXiv:2411.00027},
  year={2024}
}

@inproceedings{huang2023learning,
  title={Learning Retrieval Augmentation for Personalized Dialogue Generation},
  author={Huang, Qiushi and Fu, Shuai and Liu, Xubo and Wang, Wenwu and Ko, Tom and Zhang, Yu and Tang, Lilian},
  booktitle={Proceedings of the 2023 Conference on Empirical Methods in Natural Language Processing},
  pages={2523--2540},
  organization={Association for Computational Linguistics}
}

@article{luo2024integrating,
  title={Integrating large language models into recommendation via mutual augmentation and adaptive aggregation},
  author={Luo, Sichun and Yao, Yuxuan and He, Bowei and Shao, Wei and Xu, Jian and Huang, Yinya and Zhou, Aojun and Zhang, Xinyi and Xiao, Yuanzhang and Hou, Hanxu and others},
  journal={IEEE Journal of Selected Topics in Signal Processing},
  year={2026},
  publisher={IEEE}
}

@article{richardson2023integrating,
  title={Integrating summarization and retrieval for enhanced personalization via large language models},
  author={Richardson, Chris and Zhang, Yao and Gillespie, Kellen and Kar, Sudipta and Singh, Arshdeep and Raeesy, Zeynab and Khan, Omar Zia and Sethy, Abhinav},
  journal={arXiv preprint arXiv:2310.20081},
  year={2023}
}

@article{jiang2024mixtral,
  title={Mixtral of experts},
  author={Jiang, Albert Q and Sablayrolles, Alexandre and Roux, Antoine and Mensch, Arthur and Savary, Blanche and Bamford, Chris and Chaplot, Devendra Singh and Casas, Diego de las and Hanna, Emma Bou and Bressand, Florian and others},
  journal={arXiv preprint arXiv:2401.04088},
  year={2024}
}

@article{anil2023palm,
  title={Palm 2 technical report},
  author={Anil, Rohan and Dai, Andrew M and Firat, Orhan and Johnson, Melvin and Lepikhin, Dmitry and Passos, Alexandre and Shakeri, Siamak and Taropa, Emanuel and Bailey, Paige and Chen, Zhifeng and others},
  journal={arXiv preprint arXiv:2305.10403},
  year={2023}
}

@article{brown2020language,
  title={Language models are few-shot learners},
  author={Brown, Tom and Mann, Benjamin and Ryder, Nick and Subbiah, Melanie and Kaplan, Jared D and Dhariwal, Prafulla and Neelakantan, Arvind and Shyam, Pranav and Sastry, Girish and Askell, Amanda and others},
  journal={Advances in neural information processing systems},
  volume={33},
  pages={1877--1901},
  year={2020}
}

@article{achiam2023gpt,
  title={Gpt-4 technical report},
  author={Achiam, Josh and Adler, Steven and Agarwal, Sandhini and Ahmad, Lama and Akkaya, Ilge and Aleman, Florencia Leoni and Almeida, Diogo and Altenschmidt, Janko and Altman, Sam and Anadkat, Shyamal and others},
  journal={arXiv preprint arXiv:2303.08774},
  year={2023}
}

@article{touvron2023llama,
  title={Llama: Open and efficient foundation language models},
  author={Touvron, Hugo and Lavril, Thibaut and Izacard, Gautier and Martinet, Xavier and Lachaux, Marie-Anne and Lacroix, Timoth{\'e}e and Rozi{\`e}re, Baptiste and Goyal, Naman and Hambro, Eric and Azhar, Faisal and others},
  journal={arXiv preprint arXiv:2302.13971},
  year={2023}
}

@article{touvron2023llama2,
  title={Llama 2: Open foundation and fine-tuned chat models},
  author={Touvron, Hugo and Martin, Louis and Stone, Kevin and Albert, Peter and Almahairi, Amjad and Babaei, Yasmine and Bashlykov, Nikolay and Batra, Soumya and Bhargava, Prajjwal and Bhosale, Shruti and others},
  journal={arXiv preprint arXiv:2307.09288},
  year={2023}
}

@article{luo2024perfedrec++,
  title={Perfedrec++: Enhancing personalized federated recommendation with self-supervised pre-training},
  author={Luo, Sichun and Xiao, Yuanzhang and Zhang, Xinyi and Liu, Yang and Ding, Wenbo and Song, Linqi},
  journal={ACM Transactions on Intelligent Systems and Technology},
  volume={15},
  number={5},
  pages={1--24},
  year={2024},
  publisher={ACM New York, NY, USA}
}

@inproceedings{luo2022personalized,
  title={Personalized federated recommendation via joint representation learning, user clustering, and model adaptation},
  author={Luo, Sichun and Xiao, Yuanzhang and Song, Linqi},
  booktitle={Proceedings of the 31st ACM international conference on information \& knowledge management},
  pages={4289--4293},
  year={2022}
}

@incollection{jameson2007adaptive,
  title={Adaptive interfaces and agents},
  author={Jameson, Anthony},
  booktitle={The human-computer interaction handbook},
  pages={459--484},
  year={2007},
  publisher={CRC press}
}

@article{schiaffino2004user,
  title={User--interface agent interaction: personalization issues},
  author={Schiaffino, Silvia and Amandi, Anal{\i}a},
  journal={International Journal of Human-Computer Studies},
  volume={60},
  number={1},
  pages={129--148},
  year={2004},
  publisher={Elsevier}
}

@article{kocaballi2019personalization,
  title={The personalization of conversational agents in health care: systematic review},
  author={Kocaballi, Ahmet Baki and Berkovsky, Shlomo and Quiroz, Juan C and Laranjo, Liliana and Tong, Huong Ly and Rezazadegan, Dana and Briatore, Agustina and Coiera, Enrico},
  journal={Journal of medical Internet research},
  volume={21},
  number={11},
  pages={e15360},
  year={2019},
  publisher={JMIR Publications Toronto, Canada}
}

@article{araujo2024speaking,
  title={From speaking like a person to being personal: The effects of personalized, regular interactions with conversational agents},
  author={Araujo, Theo and Bol, Nadine},
  journal={Computers in Human Behavior: Artificial Humans},
  volume={2},
  number={1},
  pages={100030},
  year={2024},
  publisher={Elsevier}
}

@inproceedings{wangmathcoder,
  title={MathCoder: Seamless Code Integration in LLMs for Enhanced Mathematical Reasoning},
  author={Wang, Ke and Ren, Houxing and Zhou, Aojun and Lu, Zimu and Luo, Sichun and Shi, Weikang and Zhang, Renrui and Song, Linqi and Zhan, Mingjie and Li, Hongsheng},
  booktitle={The Twelfth International Conference on Learning Representations}
}

@inproceedings{zhousolving,
  title={Solving Challenging Math Word Problems Using GPT-4 Code Interpreter with Code-based Self-Verification},
  author={Zhou, Aojun and Wang, Ke and Lu, Zimu and Shi, Weikang and Luo, Sichun and Qin, Zipeng and Lu, Shaoqing and Jia, Anya and Song, Linqi and Zhan, Mingjie and others},
  booktitle={The Twelfth International Conference on Learning Representations}
}

@inproceedings{gu2023llm,
  title={Llm-based code generation method for golang compiler testing},
  author={Gu, Qiuhan},
  booktitle={Proceedings of the 31st ACM Joint European Software Engineering Conference and Symposium on the Foundations of Software Engineering},
  pages={2201--2203},
  year={2023}
}

@article{liu2024exploring,
  title={Exploring and evaluating hallucinations in llm-powered code generation},
  author={Liu, Fang and Liu, Yang and Shi, Lin and Huang, Houkun and Wang, Ruifeng and Yang, Zhen and Zhang, Li and Li, Zhongqi and Ma, Yuchi},
  journal={arXiv preprint arXiv:2404.00971},
  year={2024}
}

@inproceedings{kwon2023efficient,
  title={Efficient memory management for large language model serving with pagedattention},
  author={Kwon, Woosuk and Li, Zhuohan and Zhuang, Siyuan and Sheng, Ying and Zheng, Lianmin and Yu, Cody Hao and Gonzalez, Joseph and Zhang, Hao and Stoica, Ion},
  booktitle={Proceedings of the 29th Symposium on Operating Systems Principles},
  pages={611--626},
  year={2023}
}

@article{yang2024qwen2,
  title={Qwen2.5 technical report},
  author={Yang, An and Yang, Baosong and Zhang, Beichen and Hui, Binyuan and Zheng, Bo and Yu, Bowen and Li, Chengyuan and Liu, Dayiheng and Huang, Fei and Wei, Haoran and others},
  journal={arXiv preprint arXiv:2412.15115},
  year={2024}
}

@inproceedings{weng2023large,
  title={Large Language Models are Better Reasoners with Self-Verification},
  author={Weng, Yixuan and Zhu, Minjun and Xia, Fei and Li, Bin and He, Shizhu and Liu, Shengping and Sun, Bin and Liu, Kang and Zhao, Jun},
  booktitle={Findings of the Association for Computational Linguistics: EMNLP 2023},
  pages={2550--2575},
  year={2023}
}

@inproceedings{wangself,
  title={Self-Consistency Improves Chain of Thought Reasoning in Language Models},
  author={Wang, Xuezhi and Wei, Jason and Schuurmans, Dale and Le, Quoc V and Chi, Ed H and Narang, Sharan and Chowdhery, Aakanksha and Zhou, Denny},
  booktitle={The Eleventh International Conference on Learning Representations}
}

@article{lewis2020retrieval,
  title={Retrieval-augmented generation for knowledge-intensive nlp tasks},
  author={Lewis, Patrick and Perez, Ethan and Piktus, Aleksandra and Petroni, Fabio and Karpukhin, Vladimir and Goyal, Naman and K{\"u}ttler, Heinrich and Lewis, Mike and Yih, Wen-tau and Rockt{\"a}schel, Tim and others},
  journal={Advances in neural information processing systems},
  volume={33},
  pages={9459--9474},
  year={2020}
}

@article{luo2026relay,
  title={Relay, Don't Route: Adaptive Population Handoff for Cost-Efficient LLM-Driven Evolution},
  author={Luo, Sichun and Huang, Yi and Deng, Guanzhi and Wang, Haibo and Luo, Haochen and Li, Lei and Hu, Zefa and Feng, Junlan and Liu, Qi},
  journal={arXiv preprint arXiv:2608.05651},
  year={2026}
}

@article{luo2026seaevo,
  title={SeaEvo: Advancing Algorithm Discovery with Strategy Space Evolution},
  author={Luo, Sichun and Huang, Yi and Luo, Haochen and Liu, Fengyuan and Deng, Guanzhi and Li, Lei and Yao, Qinghua and Hu, Zefa and Feng, Junlan and Liu, Qi},
  journal={arXiv preprint arXiv:2604.24372},
  year={2026}
}

@inproceedings{luo2026alphaqt,
  title={ALPHAQT-BENCH: Diagnosing the Gap between Financial Code Generation and Quantitative Reasoning in LLMs},
  author={Luo, Sichun and Huang, Yi and Meng, Shichang and Liu, Fengyuan and Li, Mukai and Yao, Qinghua and Hu, Zefa and Feng, Junlan and Liu, Qi},
  booktitle={Findings of the Association for Computational Linguistics: ACL 2026},
  pages={2871--2886},
  year={2026}
}

@article{luo2025clarifymt,
  title={Clarifymt-bench: Benchmarking and improving multi-turn clarification for conversational large language models},
  author={Luo, Sichun and Huang, Yi and Li, Mukai and Meng, Shichang and Liu, Fengyuan and Hu, Zefa and Feng, Junlan and Liu, Qi},
  journal={arXiv preprint arXiv:2512.21120},
  year={2025}
}

@article{luo2025recranker,
  title={Recranker: Instruction tuning large language model as ranker for top-k recommendation},
  author={Luo, Sichun and He, Bowei and Zhao, Haohan and Shao, Wei and Qi, Yanlin and Huang, Yinya and Zhou, Aojun and Yao, Yuxuan and Li, Zongpeng and Xiao, Yuanzhang and others},
  journal={ACM Transactions on Information Systems},
  volume={43},
  number={5},
  pages={1--31},
  year={2025},
  publisher={ACM New York, NY}
}

\newpage

\appendix

\section{Evaluation Details}
\label{sec:eval_detail}

\textbf{Hardware and Software.} We conduct the main experiments on a server with CPU:
Intel(R) Xeon(R) Gold 6354 CPU @ 3.00GHz and GPU: NVIDIA A40 48GB using Python 3.10.14.
We use vLLM \cite{kwon2023efficient} for inference acceleration.

\textbf{LaMP Dataset Introduction}
Personalized Citation Identification (LaMP-1): This task focuses on identifying relevant citations tailored to a user's specific research interests and past citation patterns.
Personalized News Categorization (LaMP-2N): In this task, the model categorizes news articles based on user preferences, such as past reading history and topical interests.
Personalized Movie Tagging (LaMP-2M): The model assigns personalized tags to movies based on user viewing history and taste.
Personalized Product Rating (LaMP-3): This task evaluates the model's ability to predict product ratings that align with individual user preferences and prior feedback.
Personalized News Headline Generation (LaMP-4): The goal is to generate news headlines that are tailored to a user's interests, ensuring relevance and engagement.
Personalized Scholarly Title Generation (LaMP-5): This task involves generating scholarly article titles that reflect a user's research focus and academic style.
Personalized Tweet Paraphrasing (LaMP-7): The model paraphrases tweets, taking into account the user's writing style and preferred tone.


\section{Details of the Supplementary Analyses}
\label{sec:supp_detail}

The analyses in Section~\ref{sec:quant_limitations} and the backbone study in
Section~\ref{sec:backbone} follow the protocol of Section~\ref{sec:exp_setup}
(200 users, BM25 retrieval, $T=0.6$ for reasoning models). The backbone runs and
the controlled reruns used for Table~\ref{tab:divergence} were executed on
NVIDIA L20 GPUs with vLLM, and each configuration is repeated three times with
different seeds; the reported values are means over the three runs.

\textbf{Format-violation criterion.} A generation counts as valid only if the
text following the reasoning trace, after stripping surrounding whitespace,
quotation marks and Markdown emphasis, is exactly one of the task labels.
Answers that are correct but wrapped in an explanatory sentence
(e.g.\ ``The category for this article is \emph{sports}.'') therefore count as
violations, since they require task-specific post-processing.

\textbf{Backtracking markers.} We count case-insensitive whole-word occurrences
of \textit{wait}, \textit{hmm}, \textit{alternatively}, \textit{however},
\textit{actually}, \textit{but}, \textit{maybe} and the phrase
\textit{let me reconsider} inside the reasoning trace, normalised by the number
of whitespace-separated words of that trace. For \model the trace is the
concatenation of the template-guided chain and the post-intervention
continuation, so the reported rate accounts for both stages.

\textbf{Pairwise agreement.} For each task we take the three repeated runs and
average the three pairwise fractions of users for which the extracted answers
are string-identical.

\section{Prompt Template}
\label{sec:prompt}

\begin{tcolorbox}  
\label{prompt-hierarchical-thought-templates}
{\slshape 
\textbf{Handcrafted Prompt for Hierarchical Reasoning Thought Template}\\
Please follow these steps to handle user queries and ensure the output meets requirements:
1. Question Analysis: Carefully analyze the user input to clarify the intent and specific needs of the query. Extract key information such as task type (e.g., judgment, generation, summary) and constraints (e.g., format, scope).
2. User Profile Integration: Use the background information provided by the user (e.g., research field, interests, past queries) to infer preferences and personalize the content. If no specific profile is available, assume the user needs accurate and professional answers.
3. Retrieval Augmented Generation: If the external information (e.g., references, data) is referenced, prioritize using the context provided by the user to avoid irrelevant assumptions.
4. Content Generation: Based on the question analysis, user profile, and retrieved information, generate a preliminary answer that meets the requirements. Ensure the content is accurate, relevant, and logically clear.
5. Format Alignment and Checking: Verify that the output conforms to any specified format requirements (e.g., "answer only" or "no explanation"). If no format is specified, default to a concise and readable style.
6. Conciseness Optimization: Remove redundant information and avoid overthinking or irrelevant details. Ensure the answer directly addresses the core of the query.
7. Final Output: Present the final answer clearly and concisely, ensuring it meets user needs.
}
\end{tcolorbox}



\section{Limitation}
\label{sec:limitation}
Due to computational constraints, our evaluation is limited to a subset of large reasoning models (LRMs), specifically Llama, Qwen, and DeepSeek-R1-Distill series, leaving other prominent LRMs, such as OpenAI o1, untested. This restricts the generalizability of our findings across the broader spectrum of reasoning models. Additionally, our study primarily focuses on training-free methods, such as the proposed \model framework, while supervised fine tuning approaches remain underexplored. Future work could address these gaps by evaluating a wider range of models and incorporating alignment strategies to further enhance personalization performance. Lastly, the LaMP benchmark, while comprehensive, may not capture all real-world personalization scenarios, particularly those involving dynamic user interactions or multimodal data, which could limit the applicability of our findings in more complex settings.

\section{Broad Impact}
\label{sec:broad}

This work advances the personalization capabilities of large reasoning models (LRMs). By improving the ability of LRMs to generate user-specific outputs, our proposed framework can enhance user experiences in applications such as personalized education, recommendations, and conversational agents. However, these advancements also introduce potential negative societal impacts. Enhanced personalization may exacerbate privacy concerns, as models rely on detailed user profiles that could be misused if not properly safeguarded. For example, unauthorized access to user data could lead to targeted manipulation or discrimination. 

\begin{figure}[!t]
\begin{tcolorbox}[colback=wkyellow!50!white,colframe=wkyellow!80!orange,title=\textcolor{black}{(a) Question}]
\begin{small}
Generate a title for the following abstract of a paper: Recent years have witnessed the explosive growth of online social networks (OSNs). They provide powerful IT-innovations for online social activities such as organizing contacts, publishing content, and sharing interests between friends who may never meet before. As more and more people become active users of OSNs, one may ponder questions such as (1) Do OSNs indeed improve our sociability? (2) To what extent can we expand our offline social spectrum in OSNs? (3) Can we identify some interesting user behaviors in OSNs? Our work in this paper attempts to answer these interesting questions. First, we systematically validate the existence of a new Dunbar@?s number in OSNs, which is ranging from 200 to 300 empirically. To reach this, we conduct local-structure analysis as well as user-interaction analysis on extensive real-world OSNs. Second, based on this new number, we divide OSN users into two categories: the rational and the aggressive, and find that rational users intend to develop close and reciprocated relationship, whereas aggressive users have no consistent behaviors. Third, we propose a simple model to highlight the constraints of time and cognition that may affect the evolution of OSNs heavily. Finally, we discuss the potential use of our findings for viral marketing and privacy management in OSNs.
\end{small}
\end{tcolorbox}
\begin{tcolorbox}[colback=wkgreen!50!white,colframe=wkgreen!95!black, title=\textcolor{black}{(b) Answer generated by Llama 3:}]
\begin{small}
Characterizing User Behavior in Online Social Networks: A Study on Sociability, User Categories, and Evolution

\end{small}
\end{tcolorbox}

\begin{tcolorbox}[colback=wkblue!50!white,colframe=wkblue!80!blue,title=\textcolor{black}{(c) Answer generated by DeepSeek-Llama3: }]
\begin{small}
<think> Okay, the user has provided an abstract of a paper and wants me to generate a title for it. Let me read through the abstract carefully to understand the key points. The paper discusses the growth of online social networks (OSNs) and addresses three main questions about their impact on sociability, expanding the offline social spectrum, and identifying user behaviors. First, the study validates a new Dunbar number for OSNs, which is empirically found to range from 200 to 300. They did this through local-structure and user-interaction analyses on real-world OSNs. So, the title should probably mention something about the Dunbar number or the empirical findings. Next, the paper categorizes users into rational and aggressive. Rational users form close, reciprocated relationships, while aggressive users don't show consistent behavior. This suggests the title should reflect these categories or the implications of user behavior on OSNs. They also propose a model about constraints of time and cognition affecting OSN evolution. This adds another layer to the title, possibly focusing on the factors influencing the network's growth or dynamics. Furthermore, the findings are discussed in the context of viral marketing and privacy management. The title should indicate the practical applications or implications of their study. Putting this together, I need a title that encapsulates all these aspects: the empirical discovery of a new Dunbar number, the categorization of users, the proposed model, and the applications. Maybe something like  Empirical Insights into Online Social Networks: The Dunbar Number, User Behavior, and Implications for Marketing and Privacy. I think this covers all the main points without being too wordy. It starts with the methodology (empirical), then the specific finding (Dunbar number), the user behavior aspect, and ends with the applications. That should make the title comprehensive and informative.</think> \\Empirical Insights into Online Social Networks: The Dunbar Number, User Behavior, and Implications for Marketing and Privacy 

\end{small}
\end{tcolorbox}

\caption{An example of answers generated by Llama3 and DeepSeek-Llama3.}

\label{fig:appen}
\end{figure}

\end{document}